\documentclass[final]{hubsarxiv}
\graphicspath{{figures/}}

\usepackage{tabularx}
\usepackage{threeparttable}
\usepackage{listings}
\usepackage{setspace}

\lstdefinelanguage{SPARQL}{
  morekeywords={SELECT,DISTINCT,WHERE,PREFIX,OPTIONAL,UNION,FILTER,
                COUNT,GROUP,BY,ORDER,LIMIT,ASK,CONSTRUCT,DESCRIBE,
                FROM,NAMED,AS,NOT,EXISTS,MINUS,BIND,VALUES,SERVICE},
  sensitive=true,
  morecomment=[l]{\#},
  morestring=[b]",
}
\title{Build2SPARQL: A Large-Scale Text-to-SPARQL Benchmark Dataset for Building Knowledge Graph Querying}
\shorttitle{Build2SPARQL: a text-to-SPARQL benchmark for building knowledge graphs}
\shortauthors{Jung}
\author{Wooyoung Jung\affmark{a,}\corrmark\orcid{0000-0003-2746-4983}}
\affil{\affmark{a}Department of Civil and Architectural Engineering and Mechanics, University of Arizona, Tucson, AZ, USA}
\corresponding{wooyoung@arizona.edu}
\paperno{2026-007}
\paperdate{September 2026}
\versionnote{v1 (2026-09-22): first release. v2 (2026-11-05): re-set in the HUBS Lab paper layout; content unchanged.}
\license{CC BY-NC-ND 4.0}
\keywords{building knowledge graphs; Brick ontology; ASHRAE 223P; SPARQL; text-to-SPARQL; benchmark dataset; large language models; knowledge graph grounding; human validation}
\abstract{Building automation systems are increasingly represented as semantic knowledge graphs (KGs) using
ontologies such as Brick and ASHRAE~223P, creating a machine-readable substrate for
artificial-intelligence applications. One promising application is translating natural-language
questions into SPARQL (text-to-SPARQL), which would let building operators query these graphs
through language agents, but progress is limited by the scarcity of large natural-language/SPARQL
benchmarks. This paper presents \textbf{Build2SPARQL}, a large-scale benchmark for building KGs
generated by a KG-grounded pipeline: SPARQL queries are produced and validated entirely by
graph-traversal code, while large language models generate only the natural-language questions,
keeping query correctness independent of model behavior. The pipeline mines six query-pattern
families---linear chains, branching, UNION, aggregation, OPTIONAL, and attribute-filtered---and
phrases each query across five vocabulary registers. Applied to 201 building KGs (180 Brick, 21
ASHRAE~223P), it yields 6{,}136 executable SPARQL queries and 30{,}680 questions. A two-rater human
validation of 300 questions found 98.8\% semantic fidelity, 98.8\% naturalness, and 84.0\%
operational plausibility. A retrieval-augmented evaluation across three open-weight language models
raised exact-match accuracy from 0.2--20\% (zero-shot) to 56--65\% (three-shot retrieved).}

\begin{document}
\maketitle

\section*{Practical Applications}

For building operators, energy managers, and commissioning engineers, the value of this work is a
foundation for software that answers questions about a building in plain language. Without
learning a formal query language or a building's data schema, a practitioner could ask how
equipment is connected---for example, which air handler serves a given zone, or how many
variable-air-volume boxes each air handler feeds---and receive an answer drawn directly from the
building's own model. The dataset gives software developers a dependable basis for training and
testing such assistants: each question is paired with a query already verified to run correctly
against its source building model, so tools can be evaluated against trustworthy ground truth without
labeling examples by hand one building at a time. Practical applications include
natural-language front-ends for building analytics platforms, fault-detection and commissioning aids
that surface equipment relationships on demand, and assistants that generate grounded,
building-specific energy-saving recommendations. As these tools mature, facility teams gain a
faster, lower-expertise path to interrogating complex building systems, and the wider research
community gains a common yardstick for measuring progress. The dataset is publicly
released.

\section{Introduction}
\label{sec:intro}

Commercial and institutional buildings are among the most complex cyber-physical systems (CPSs) in modern 
society, integrating numerous sensors, actuators, and control loops across heating, ventilation, and 
air-conditioning (HVAC), lighting, metering, and access subsystems. Historically, the metadata describing 
these systems -- equipment identities, spatial containment, fluid flow paths, and sensing relationships -- has 
been locked in proprietary naming conventions that differ across vendors, campuses, and building management 
systems, making cross-building analytics and portable applications difficult to develop and maintain~\cite{fierro2019mortar}. 
The Brick ontology~\cite{balaji2018brick} addresses this fragmentation by providing an open, graph-based 
schema for building metadata: equipment classes, their relationships, and the physical and logical 
structure of a building are represented as Resource Description Framework (RDF) triples, enabling
querying with the SPARQL Protocol and RDF Query Language (SPARQL)~\cite{harris2013sparql11} over a unified, vendor-neutral model. Since its introduction, Brick has become the reference schema
for a growing ecosystem of portable building analytics applications.

As semantic building model adoption scales, so does the need for natural-language (NL) interfaces: building operators and
energy managers are domain experts, not graph query authors, and the gap between what they want to know
and what they can express in SPARQL is a practical barrier to realizing the full value of semantic
building models in workflows ranging from fault detection and energy auditing to systems commissioning. This need is sharpened by the rapid emergence of large language model (LLM)-based conversational agents and assistants for building operation and energy management~\cite{jung2026promptstrategy,jung2026simusers,jung2026domainknow,jung2026mailearn,jung2026hema}, which let operators and occupants interact with building systems in natural language but depend on structured, queryable access to each building's data to ground their outputs. The natural-language interface to these graphs is SPARQL query generation: a building operator types
a question in English, and a model translates it into an executable SPARQL query against the building's knowledge 
graphs (KGs). However, text-to-SPARQL in the building domain is particularly challenging: building KGs can be 
large (thousands of triples), highly heterogeneous across buildings, and structured around a complex schema with 
hundreds of classes and properties. Moreover, the same information can be expressed in many different ways, from 
ontology-literal to operational English, and the model must learn to capture this linguistic diversity while 
grounding it in the specific structure of a target building's KG.

As research efforts on text-to-SPARQL for buildings have emerged recently, the lack of a large,
high-quality benchmark dataset of NL/SPARQL question-answer (QA) pairs has become a critical
bottleneck. Such datasets are essential for training and evaluating text-to-SPARQL models,
developing retrieval-augmented building query systems, and assessing KG completeness -- yet no
resource for the building domain combines broad query-pattern coverage with many source buildings.
The existing benchmark, BuildingQA~\cite{mulayim2025buildingqa}, contains 27 SPARQL queries and 188
natural-language phrasings across four buildings. However, the dataset is too small to train a
specialized model, support robust evaluation, or characterize model behavior across building types
and query complexities (more details in the Related Work section).

To address this gap, \textbf{Build2SPARQL} is presented, a large-scale NL/SPARQL benchmark dataset
for Brick-schema building KGs. The corpus comprises 6{,}136 executable SPARQL queries
and 30{,}680 NL questions derived from 201 building KGs (81 real, 120 synthetic),
covering six query-pattern families defined over the static structure of named-node KGs: linear
chains (1--5 hops), branching Y-shape queries, UNION disjunctions, COUNT aggregations, OPTIONAL
extensions, and attribute-filtered queries. These families address topological and attribute
information needs. Correctness
is guaranteed by design: SPARQL queries are produced entirely by graph-traversal code that mines
patterns from and validates every query against the source building graph, while large language
models (LLMs) are responsible only for NL generation downstream of a verified query. Each validated
query is paired with five NL questions spanning a register spectrum from
ontology-literal technical language to plain operational English, conditioned on register-specific
prompts that produce vocabulary and structural variation across the corpus
(see the Methodology section).

This research makes the following contributions.
\textbf{Build2SPARQL corpus:} 6{,}136 executable SPARQL queries and 30{,}680 NL questions
        across 201 building KGs, with six query-pattern families and five
        register-differentiated questions per query.
\textbf{Quality evidence:} Empirical quality metrics including lexical diversity (mean
        self-similarity 0.170 across five register levels) and register-level structural
        validation, demonstrating that corpus-scale KG-grounded generation does not sacrifice
        question quality.
\textbf{Baseline evaluation:} Zero-shot and retrieval-augmented few-shot text-to-SPARQL
        results on a held-out test split across three open-weight LLMs, establishing a
        reproducible performance baseline for future work.
\textbf{KG-grounded generation pipeline:} A documented pipeline for Brick
        KGs that separates structural correctness (graph-traversal and
        triple-walking validation) from linguistic generation (LLMs), guaranteeing every
        emitted query is executable against the source graph.

\section{Related Work}
\label{sec:related}

\subsection{Benchmarks for Natural-Language/SPARQL: General Domain}

Benchmark datasets for NL/SPARQL question answering have grown substantially in scale
and structural complexity on general-purpose KGs~\cite{lan2021kbqasurvey}. LC-QuAD~\cite{trivedi2017lcquad}
introduced 5{,}000 question--SPARQL pairs over DBpedia covering multi-hop and count patterns;
LC-QuAD~2.0~\cite{dubey2019lcquad2} expanded this to 30{,}000 pairs over both Wikidata and
DBpedia with richer compositional templates. The QALD (Question Answering over Linked Data) shared-task 
series~\cite{usbeck2017qald} has run ten benchmarking rounds with progressively larger and more varied 
question sets, targeting DBpedia, Wikidata, and MusicBrainz. In the structurally analogous natural-language-to-SQL
(Structured Query Language) domain, Spider~\cite{yu2018spider} established a cross-domain benchmark 
of 10{,}181 question--SQL pairs across 200 databases; its successor BIRD~\cite{li2023bird} scales 
to 12{,}751 pairs over 95 real-world databases and reported only 54.89\% accuracy for GPT-4 at its introduction, underscoring that grounding
natural-language queries in schema-constrained structured data remains an open challenge even for
frontier models~\cite{rajkumar2022evaluating}.

However, all of these resources target encyclopedic, entity-rich KGs whose schema and
population differ fundamentally from the topology- and equipment-centric structure of building
KGs. Building graphs are \emph{named-node} graphs: data values are class instances and
equipment identifiers encoded as Internationalized Resource Identifiers (IRIs), the schema covers a specialized domain with hundreds of
classes and properties, and queries navigate equipment hierarchies rather than encyclopedic entity
relations. These structural differences mean that models trained on general-domain benchmarks do not
transfer to building KGs without domain-specific training data.

\subsection{Building-Domain Benchmarks and QA Data Construction}

BuildingQA~\cite{mulayim2025buildingqa} is the only prior NL/SPARQL benchmark targeting building
KGs. Its construction followed a multi-stage process: domain experts in the building
science community contributed 27 SPARQL queries they had developed for their own applications,
together with the corresponding building models; GPT-4o-mini then augmented each query with five
natural-language phrasings ranging from literal ontology-term mappings to progressively more
abstract operational inquiries, producing a final curated set of 188 questions across four buildings
(three modeled in Brick, one in American Society of Heating, Refrigerating and Air-Conditioning
Engineers (ASHRAE)~223P). This expert-grounded construction gives strong
ecological validity and intentionally preserves question ambiguity: a lexical gap analysis found
only 9--13\% direct term overlap between questions and KG terms, significantly wider than
general-domain benchmarks (11--58\%), reflecting the tendency of building operators to describe
entities by function rather than ontology class name. However, at 27 queries across four buildings,
BuildingQA may not support model training or robust evaluation.

Other building-specific text-to-SPARQL systems highlight the same bottleneck from different angles.
DA-KGQA~\cite{mulayim2025zero} evaluates a domain-adapted KGQA method on a small hand-crafted set of
19 questions over a single Mortar building, explicitly acknowledging that a larger set ``would be
prohibitively expensive'' to produce. OntoSage~\cite{devmane2026ontosage} instead fine-tunes a T5
model on roughly 120{,}000 synthetically generated NL/SPARQL pairs -- validated against a live
endpoint and evaluated on a 50-pair test set across three building deployments -- but its released
pairs are Brick-only, and its aggregation and temporal reasoning are handled by downstream analytics
microservices rather than by SPARQL structural patterns; it names the underlying data gap directly:
``no domain-specific LLMs for the smart buildings domain are available for text-to-SPARQL
translation in QA.'' BuildingGPT~\cite{li2025buildinggpt} and BuildingGPT2~\cite{li2026buildinggpt2}
scale up by generating pairs automatically: both enumerate the classes, entities, and relationships
of 45 Brick models from the Mortar testbed and instantiate a fixed set of six SPARQL templates --
retrieving all instances of a class, counting instances of a class, looking up an entity's type,
traversing one- or two-step relationships, and a class-filtered relationship retrieval -- then
annotate each query with a single LLM-generated question, yielding 18{,}555 question--query pairs
(BuildingGPT2 fine-tunes on 40 of the 45 buildings, 17{,}019 pairs, holding out five). This attains
large scale, but the fixed template set is confined to linear one- to two-hop retrieval, flat
instance counting, and class-type filtering over a single ontology (Brick), with one phrasing per
query. Branching (multi-attribute) joins, \texttt{UNION} disjunctions, \texttt{OPTIONAL} extensions,
grouped aggregation (\texttt{COUNT} with \texttt{GROUP BY}), unit/quantity attribute filters (e.g.,
sensors reporting in parts per million (ppm)), and multi-namespace traversal (e.g., ASHRAE~223P) are
out of scope. Table~\ref{tab:qa_landscape} maps this landscape. The top-right cell (large scale with
broad pattern coverage) marks the combination that Build2SPARQL targets.

\begin{table}[tbp]
\centering
\small
\caption{NL/SPARQL dataset landscape for building KGs.}
\label{tab:qa_landscape}
\begin{spacing}{1.0}
\begin{tabular}{l>{\centering\arraybackslash}p{5cm}>{\centering\arraybackslash}p{5cm}}
\toprule
 & \textbf{Small scale} & \textbf{Large scale} \\
\midrule
\textbf{Broad patterns} &
  BuildingQA~\cite{mulayim2025buildingqa} &
  -- \\[6pt]
\textbf{Narrow patterns} &
  DA-KGQA~\cite{mulayim2025zero} &
  BuildingGPT~\cite{li2025buildinggpt},
  BuildingGPT2~\cite{li2026buildinggpt2}, OntoSage~\cite{devmane2026ontosage} \\
\bottomrule
\end{tabular}
\end{spacing}
\end{table}

\subsection{Text-to-SPARQL Methods and Automated Generation}

Methods for translating NL to SPARQL have advanced through three generations.
\textit{Template-based systems} such as TBSL (Template-Based SPARQL Lookup)~\cite{unger2012tbsl} and
SINA (Scalable Interoperable Next-generation Ontology-based Question Answering)~\cite{shekarpour2015sina}
map utterances to pre-defined SPARQL skeletons and fill entity slots
via entity linking; they offer strong correctness guarantees by construction but cannot express
queries whose structure falls outside the template inventory. \textit{Neural sequence-to-sequence
models}~\cite{dong2016language}, including attention-based encoder--decoders and fine-tuned
transformers, improved structural coverage at the cost of executability: models frequently generate
syntactically plausible but semantically invalid SPARQL referencing non-existent predicates or
entity types. \textit{LLM-based approaches} using few-shot in-context prompting and
chain-of-thought reasoning~\cite{wei2022cot,kgcot2023} have pushed accuracy further; a maximum F1 of
$\approx$38\% on BuildingQA under few-shot prompting is reported~\cite{mulayim2025buildingqa},
establishing the performance ceiling for building-domain SPARQL generation and
underscoring that training data scale is the binding constraint.

Hand-authoring benchmark data at this scale is infeasible; moreover, all four building-domain
systems above draw their corpora primarily from the Mortar testbed~\cite{fierro2019mortar}, a
curated collection of North American institutional buildings, leaving generalization to other
building types, ontologies, and naming conventions largely unvalidated. Automated generation
from a diverse building corpus is therefore the necessary path. Prior
automated approaches each trade away one of the two properties that make a benchmark useful.
\textit{Template-based generation}~\cite{zou2014natural} instantiates \emph{pre-defined} pattern
skeletons -- fixed by the author before any graph is seen -- with entity slots drawn from the KG,
then fills NL surface forms by string substitution; the resulting pairs are
executable but structurally and lexically rigid, confined to whatever query shapes the author
anticipated, a limitation directly illustrated by the narrow six-type inventory in BuildingGPT2's
training corpus. \textit{Fully LLM-generated approaches} reverse the trade-off: a language model
generates both question and SPARQL without access to the target graph, producing linguistic variety
at the cost of executability -- the model cannot verify that emitted predicates exist, that path
sequences are connected, or that query results are non-empty. \textit{KG-to-text generation},
represented by the WebNLG challenge~\cite{gardent2017webnlg}, produces fluent text from RDF
triples but targets the verbalization problem rather than SPARQL construction and does not produce
paired executable queries.

Build2SPARQL introduces a fourth strategy, \textbf{KG-grounded generation}, that differs from
template-based generation in two critical respects. First, the SPARQL pattern inventory is not
pre-specified: structural patterns are \emph{discovered} by graph-traversal code conducting breadth-first
search over the target KG, so the inventory is bounded only by what the graph actually contains rather
than what an author anticipated. Second, NL questions are generated by LLMs across five
register-specific prompts rather than by surface-form string substitution, yielding vocabulary and
syntactic diversity that rigid templates cannot provide. The KG serves as the correctness
oracle: LLMs enter the pipeline only \emph{after} a query has been validated against the source graph,
ensuring they cannot introduce structural errors into the ground-truth SPARQL. The present corpus focuses 
on Brick-schema buildings; extension to other building ontologies
is a direction for future work discussed in the Conclusion.
The full methodology is described in the Methodology section.

\section{Methodology}
\label{sec:methodology}

Build2SPARQL transforms a building KG into a paired NL/SPARQL benchmark through five sequential stages: 
pattern mining, executable validation, semantic filtering, query refinement, and natural-language question 
generation. Figure~1 illustrates the data flow.

\begin{figure}[tbp]
\centering
\includegraphics[width=0.52\linewidth]{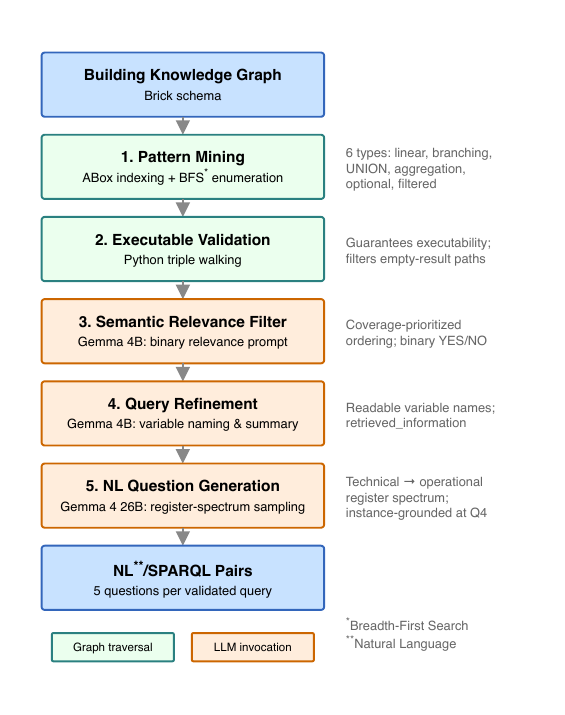}
\caption{Build2SPARQL five-stage pipeline.}
\label{fig:pipeline}
\end{figure}

The architecture enforces a strict separation between structural correctness and linguistic
generation. The six query-pattern families, all defined over the static structure of named-node
building KGs, are produced entirely in Stage~1 by graph-traversal code that 
inspects the actual graph rather than assuming a fixed schema. Stages~3--5 invoke LLMs, but only after 
every candidate query has been validated against the source graph. This ordering guarantees that no 
LLM output can introduce structural errors into the ground-truth SPARQL.

\subsection{SPARQL Pattern Mining by KG-Grounded Breadth-First Search (BFS)}
\label{sec:mining}

The six pattern types are the query families that the pipeline implements over the \emph{static
structure} of a named-node KG. Each corresponds to a core operator of relational algebra restricted
to graph-pattern matching~\cite{perez2009sparql}: selection (linear chains), conjunctive join
(branching), disjunctive join (UNION), aggregation (COUNT/GROUP BY), outer join (OPTIONAL), and
attribute filtering. The scope is deliberately bounded to what triple-walking can validate against a
static graph. Which instances of each type are generated depends on what
the target graph actually contains: the pattern inventory is discovered at runtime rather than
pre-specified. All six types are fully implemented in the pipeline; which appear in the output
depends on whether the target graph contains the structural co-occurrences each type requires.
In the current Brick corpus, linear and branching patterns dominate because the remaining four
types require co-occurrence conditions -- shared anchor-terminal pairs (UNION), sufficient
group-count variance (aggregation), partially-present properties (optional), and
measurement-type attributes (attribute-filtered) -- that the Brick buildings in this corpus
meet less frequently. The per-type distribution is reported in the Dataset section.

Pattern mining begins with \emph{ABox indexing}: a SPARQL introspection query scans the building
graph for all $(s_{\text{class}},\,p,\,o_{\text{class}})$ class-level triples and records the
instance count of each. All subsequent pattern enumeration operates over this class-level summary,
deferring raw graph access to the validation stage.

\textbf{Linear chains} cover selective retrieval questions of the form ``which equipment X has
property Y?''\ -- the primary selective retrieval pattern in building KGs. The class-level index is
traversed by breadth-first search to enumerate all acyclic connecting paths to a configurable depth. 
On dense graphs, shallow 1-hop patterns would otherwise exhaust the total pattern budget; a per-depth
cap ensures that multi-hop paths remain represented in the output.

\textbf{Branching (Y-shape) paths} enable conjunctive questions about instances satisfying two
simultaneous relationships -- ``which VAV (Variable Air Volume) units serve an HVAC zone and also have a temperature
sensor?''\ A branching path is formed from any two linear paths that share an anchor class but
diverge at different properties toward distinct terminal classes, ensuring the join expresses a
genuine two-attribute constraint rather than a degenerate self-join.

\textbf{UNION paths} capture disjunctive retrieval questions in which the same entity type is
reachable via alternative traversal routes -- ``find equipment accessible via either path A or
path B.''\ They pair linear paths sharing both anchor and terminal class but differing in the
connecting property, the disjunctive counterpart of the branching conjunctive join.

\textbf{Aggregation paths} generate ``how many X does each Y have?'' questions by wrapping linear
paths as grouped COUNT queries. To avoid skewing the corpus toward equipment-dense system types,
only the single highest-instance-count path is aggregated per anchor--terminal class pair.

\textbf{Optional paths} surface questions about partial attribute presence: ``which equipment has
this property, and which does not?''\ A candidate outgoing property at a terminal node qualifies
as an OPTIONAL extension when it appears on some but not all terminal instances -- present often
enough to be non-trivial, yet absent often enough that the optional semantics are meaningful
rather than tautological.

\textbf{Attribute-filtered paths} narrow a retrieval to instances carrying a specific measurement
type -- ``which sensors measure CO\textsubscript{2} concentration?''\ Terminal nodes are extended
with measurement-type predicates drawn from the target ontology; each distinct attribute value
appearing above the instance threshold becomes a separate filtered variant of the base linear path.
This pattern requires quantifiable properties annotated with unit or quantity-kind values (e.g.,
QUDT (Quantities, Units, Dimensions, and Data Types) \texttt{hasQuantityKind}, \texttt{hasUnit}). In the current corpus these annotations are
present only in buildings using the ASHRAE~223P (s223) schema; Brick's named-node structure carries
no such literal annotations, so attribute-filtered queries arise exclusively from the s223 buildings
in the corpus. The pattern is fully implemented and would generate queries from any
ontology that carries quantifiable property annotations.

To filter near-degenerate patterns, each candidate query must return a minimum number of result
rows at validation (described below), with the threshold scaling by building size:
1~row for fewer than 50 instances, 2 for 50--150, and 3 for larger graphs.

\subsection{Executable Validation by Triple Walking}
\label{sec:triple_walk}

Build2SPARQL validates each candidate query through direct Python triple-walking: the graph is
traversed hop-by-hop along the pattern's property path, collecting matching instance tuples.
Any pattern returning no rows is discarded. The matched tuples also serve as sample results
passed to the NL generation stage (described below), where they ground
instance-specific questions in verbatim equipment identifiers drawn from the actual building.

\subsection{Semantic Relevance Filtering with Coverage Prioritization}
\label{sec:filter}

Structural validity alone does not guarantee practical relevance. This stage uses Gemma~4~4B
with a binary prompt asking whether the pattern describes information of practical relevance
to building operation. Patterns receiving a \texttt{NO} response are discarded. The prompt
instructs the model: \textit{``Answer YES if this pattern would help answer a real question
about the building, or NO if it is trivially obvious, too generic, or not actionable.
Respond with exactly one line: YES or NO followed by a dash and one short sentence
explaining why.''}

Before filtering, patterns are ranked by a four-dimensional coverage score that quantifies how
underrepresented each pattern is in the current building's diversity state. The four dimensions are:
\emph{system type} (HVAC, lighting, metering, and others), \emph{ontology class} (the anchor
class), \emph{complexity} (hop-count bucket), and \emph{pattern type} (linear, branching, UNION,
aggregation, optional, filtered). Sorting by rarest-cell-first ensures that the filter stage
evaluates the most coverage-expanding patterns first and does not preferentially discard
underrepresented categories when budget limits are reached.

\subsection{Query Refinement and Summarization}
\label{sec:writer}

Each pattern passing the relevance filter is submitted to the same model (Gemma~4~4B) with a
prompt that asks the model to (a)~rename SPARQL variables from short class-name tokens to readable
multi-word identifiers and (b)~produce a one-sentence \texttt{retrieved\_information} summary
describing what the query returns. The model receives the building description (ontology type,
equipment class inventory) as context but does not modify the query's structural template: triple
patterns, grouping clauses, and \texttt{OPTIONAL} and \texttt{UNION} blocks are preserved exactly.
A final execution check confirms the rewritten query remains valid.

\subsection{Multi-Register Natural-Language Question Generation}
\label{sec:nlgen}

Five natural-language questions are generated per validated query using Gemma~4~31B (\texttt{gemma4:31b}).
Rather than producing five paraphrases of a single utterance, the pipeline targets five points on a 
vocabulary register spectrum -- from graph-structure-aware technical language to plain operational English -- by 
conditioning the language model on five register-specific system prompts.

Five vocabulary levels are defined by concrete behavioral instructions that span from schema-literal to
schema-free language.
\textbf{Q1~(Technical)} uses ontology class names and property identifiers verbatim, as they appear in the
SPARQL query.
\textbf{Q2~(Systems-engineering)} paraphrases ontology class names into building domain engineering
terminology (e.g., ``supply fan start/stop control'' for \texttt{brick:Start\_Stop\_Command}), without
anchoring to a specific named equipment instance.
\textbf{Q3~(Operational)} describes the information need in plain operational terms, removing both ontology
vocabulary and domain class names but still referring to equipment categories generically.
\textbf{Q4~(Instance-grounded)} embeds a specific named equipment identifier drawn verbatim from the
query's validation-walk results (e.g., \texttt{SupplyFan\_AHU01}), framing the question as a practical
operational task around that concrete object.
\textbf{Q5~(Plain English)} paraphrases the information need in plain, accessible language,
using no ontology terms, equipment identifiers, or engineering vocabulary, targeting any
building occupant without knowledge of the underlying schema.
Their collective function is to increase phrasing variety across the corpus; the five levels are not
intended as categorically separable question types. Table~\ref{tab:registers} lists the defining
instruction for each register point as it appears in the system prompt.

\begin{table}[tbp]
\centering
\small
\caption{Register-specific prompt directives for NL question generation.}
\label{tab:registers}
\begin{spacing}{1.0}
\begin{tabular}{clp{10cm}}
\toprule
Q & Register label & Key instruction \\
\midrule
1 & Technical &
  Name the equipment types and relationships in the query directly. Use ontology class names
  and property identifiers where natural. \\
2 & Systems-engineering &
  Use building domain terminology. Paraphrase ontology class names into engineering language
  but keep them recognizable as specific equipment types. \\
3 & Operational &
  Describe what information is needed without naming specific ontology classes or data model
  terms. \\
4 & Instance-grounded &
  When named entity examples are listed, use one verbatim in the question
  (e.g.\ ``Is \texttt{AHU-3} reporting\ldots'' rather than ``Is the AHU reporting\ldots'',
  where AHU denotes an Air Handling Unit).
  Describe a practical task or check, not a data query. \\
5 & Plain English &
  Write a plain English question using no ontology terms, no SPARQL vocabulary, no class
  names, and no equipment identifiers. Paraphrase the information need in everyday language
  accessible to any building occupant. \\
\bottomrule
\end{tabular}
\end{spacing}
\end{table}

The generation prompt for each register point injects (i)~the building description, (ii)~the
written SPARQL query and its \texttt{retrieved\_information} summary, (iii)~sample result rows
from the validation walk (named equipment identifiers are withheld for Q5 to prevent tag leakage
into plain-English questions), and (iv)~all previously generated questions for that query with an
instruction to use a different sentence structure and opening word. This anti-repetition
directive suppresses the tendency to default to identical sentence frames.

\subsection{Generation Configuration and Reproducibility}
\label{sec:repro}

Table~\ref{tab:genconfig} lists the models and decoding parameters for the three LLM stages. All
stages run open-weight Gemma~4 models served locally through Ollama with no fine-tuning, so
generation is reproducible without external application programming interface (API)
dependencies. The binary semantic filter and the
constrained query-rewriting stage use the 4B variant (\texttt{gemma4:e4b}); question generation uses
the larger 31B variant (\texttt{gemma4:31b}), which a model-selection comparison found produced more
accurate operational (Q4) and plain-English (Q5) phrasings. Question generation additionally injects
a pattern-type hint that steers question form for non-linear queries (e.g., disjunction phrasing for
UNION, presence/absence phrasing for OPTIONAL); for Q4, one to three named instance identifiers
extracted from the validation-walk results are supplied to the model as explicit examples. 
The per-register prompt directives are listed in Table~\ref{tab:registers} and the models and
decoding parameters in Table~\ref{tab:genconfig}, fully specifying the generation configuration.

\begin{table}[tbp]
\centering
\small
\caption{LLM generation configuration. All stages use open-weight Gemma~4 models served locally via
Ollama with no fine-tuning.}
\label{tab:genconfig}
\begin{tabular}{lccc}
\toprule
Pipeline stage & Model & Temperature & Max output tokens \\
\midrule
Semantic filtering       & Gemma~4~4B (\texttt{gemma4:e4b})  & 0.3 & 512 \\
Query refinement         & Gemma~4~4B (\texttt{gemma4:e4b})  & 0.3 & 800 \\
NL question generation   & Gemma~4~31B (\texttt{gemma4:31b}) & 0.7 & 150 \\
\bottomrule
\end{tabular}
\end{table}

\section{Build2SPARQL Dataset}
\label{sec:dataset}

Build2SPARQL has four defining properties: (1)~every SPARQL query is guaranteed executable by
KG-grounded triple-walking against its source graph; (2)~the corpus spans 201 source buildings, real
and synthetic, and 6{,}136 queries; (3)~each query is paired with five NL questions spanning a
vocabulary register spectrum from ontology-literal to plain English; and (4)~the dataset covers
multiple functional categories, enabling cross-building benchmarking.

\subsection{Corpus Overview}

The Build2SPARQL corpus is produced from 201 building KGs spanning two semantic schemas: 180 use
the Brick schema and 21 use ASHRAE~223P (s223). The 81 real buildings comprise 47 Brick buildings
from the Mortar open testbed~\cite{fierro2019mortar}, 13 curated Brick reference models covering
system configurations underrepresented in the Mortar set (metering hierarchies, air-quality sensor
arrays, and chiller-plant topologies), and 21 ASHRAE~223P example models---the standard's published
reference buildings, including the ASHRAE Guideline~36 control sequences and example models from
the National Institute of Standards and Technology (NIST), Pacific Northwest National Laboratory
(PNNL), Lawrence Berkeley National Laboratory (LBNL), and National Renewable Energy Laboratory
(NREL). The Mortar buildings are anonymised North American facilities spanning
diverse functions, including large commercial offices, university buildings, and mixed-use
institutional complexes. The ASHRAE~223P models are the source of the attribute-filtered queries,
since Brick's named-node structure carries no unit or quantity-kind annotations.

The 120 synthetic buildings are Brick graphs produced by the BuildGraph procedural
generator~\cite{buildgraph2025}, covering 8 U.S. Department of Energy (DOE) commercial prototype
archetypes (small office, medium office, large office, primary school, hospital, hotel, strip mall,
and outpatient healthcare) across three construction vintages (pre-1980, 2004, 2013), with five
instances per type-vintage cell. This selection extends the corpus with labeled structural diversity
across a representative range of DOE prototype types and energy-code eras while complementing the
organic heterogeneity of the real buildings.

Table~\ref{tab:corpus_overview} summarizes the corpus. Building KGs vary substantially in size,
from compact reference models with as few as 4 distinct equipment instances to large facility
graphs with up to 932 instances across the 201 buildings. This range ensures
that the benchmark tests retrieval over both small, well-structured graphs and large, heterogeneous
ones. The pipeline enforces a 45-minute processing limit per building to prevent individual graphs
from blocking the full corpus run; two unusually sparse Mortar buildings produced fewer than five
valid queries and are retained with their actual counts, reflecting genuine graph sparsity rather
than a pipeline limitation.

\begin{table}[h!]
\centering
\small
\caption{Build2SPARQL corpus overview.}
\label{tab:corpus_overview}
\begin{tabular}{lr}
\toprule
Metric & Count \\
\midrule
Buildings & 201 \\
\quad Real & 81 \\
\qquad Mortar testbed (Brick) & 47 \\
\qquad Brick reference models & 13 \\
\qquad ASHRAE~223P example models & 21 \\
\quad Synthetic (BuildGraph, Brick) & 120 \\
\quad Schema (Brick / ASHRAE~223P) & 180 / 21 \\
Queries & 6{,}136 \\
\quad Mean per building & 30.5 \\
NL questions (5 per query) & 30{,}680 \\
Distinct ontology terms & 232 \\
Mean Basic Graph Pattern (BGP) triples & 3.9 \\
\bottomrule
\end{tabular}
\end{table}

Real buildings show wide variability in query counts, reflecting genuine heterogeneity in graph
density and system completeness. BuildGraph buildings are more tightly clustered (408--460 queries
per type) because their regular prototype structure saturates the same pattern classes consistently
across instances.

All 6{,}136 queries use \texttt{SELECT DISTINCT} and 98\% employ \texttt{rdf:type/rdfs:subClassOf*}
property paths for subclass-aware class retrieval. Both reflect the structure of the corpus's Brick
graphs rather than pipeline choices. \texttt{SELECT DISTINCT} is necessary because multi-hop traversals can revisit
the same terminal node via multiple intermediate paths. The \texttt{rdf:type/rdfs:subClassOf*}
path is required because Brick encodes equipment as instances of specific leaf classes
(e.g., \texttt{brick:VAV\_Box}) that are subclasses of the query anchor class (\texttt{brick:VAV});
without path traversal, queries return only the fraction of instances typed at exactly the anchor
level, producing semantically incomplete results. Listing~\ref{lst:sparql_example} shows a
representative corpus query in full.

\begin{spacing}{1.0}
\begin{lstlisting}[label=lst:sparql_example,
  caption={A representative corpus query.}]
PREFIX brick: <https://brickschema.org/schema/Brick#>
PREFIX rdf:   <http://www.w3.org/1999/02/22-rdf-syntax-ns#>
PREFIX rdfs:  <http://www.w3.org/2000/01/rdf-schema#>

SELECT DISTINCT ?ahu ?vav WHERE {
  ?ahu rdf:type/rdfs:subClassOf*  brick:Air_Handler_Unit .
  ?ahu brick:feeds                ?vav .
  ?vav rdf:type/rdfs:subClassOf*  brick:VAV .
}
\end{lstlisting}
\end{spacing}

Every query record carries two complementary text fields: a \texttt{description} field providing
the building-context paragraph fed to the LLM during NL generation (equipment class inventory and
key inter-class relationships), and a \texttt{retrieved\_information} field, a one-sentence
plain-language summary of what the query retrieves, which supports retrieval-augmented prompting
use cases described in the Discussion section. Every record contains exactly five questions, one per register point;
there are no partial records.

\subsection{Pattern-Type Distribution}
\label{sec:pattern_dist}

Table~\ref{tab:pattern_dist} shows the per-pattern-type breakdown for the current corpus.
Linear chains dominate at 81.6\%, reflecting the topology of the eight DOE prototype archetypes:
their regular, repeating equipment hierarchies produce dense chains but relatively few
multi-attribute co-occurrences. The remaining five non-linear types each require structural
co-occurrences that are met less frequently: two distinct dense outgoing relationship types per
anchor class (branching), shared anchor--terminal class pairs reachable via different property
paths (UNION), sufficient group-count variance across anchor instances (aggregation),
partially-present outgoing properties at terminal nodes (optional), and measurement-type
attribute annotations at terminal nodes (attribute-filtered).

Real buildings produce a higher per-building branching rate than synthetic buildings, as their
equipment classes more often participate simultaneously in multiple dense relationship types,
creating the two-branch co-occurrence that branching queries require. The 81.6\% linear share
reflects building topology rather than a pipeline bias: the mining stage implements all six pattern
types, but the Brick corpus satisfies the non-linear co-occurrence conditions less frequently
(Methodology section). The skew is nonetheless a genuine consideration for downstream use, since a
model trained or evaluated on the raw distribution is dominated by linear-chain behavior, and a
single aggregate accuracy figure understates the difficulty of the rarer structures -- in the
NL-to-SPARQL evaluation, UNION queries score far below linear chains under
identical conditions. Three practices are therefore recommended, all supported by the
\texttt{pattern\_type} field on every record: (i)~report per-pattern metrics rather than a single
aggregate, as done here; (ii)~oversample or loss-reweight the rarer types when training, to
counteract the bias toward linear retrieval; and (iii)~for difficulty-sensitive evaluation, use the
1{,}127 non-linear queries as a targeted challenge set, or draw a fully pattern-balanced subset by
capping each type at the size of the rarest (optional, 41 queries; 246 queries balanced).

\begin{table}[tbp]
\centering
\small
\caption{Query pattern-type distribution.}
\label{tab:pattern_dist}
\begin{tabular}{lrr}
\toprule
Pattern type & Queries & Share (\%) \\
\midrule
Linear chain         & 5{,}009 & 81.6 \\
Branching (Y-shape)  &   815   & 13.3 \\
Aggregation (COUNT)  &   120   &  2.0 \\
Attribute-filtered   &   100   &  1.6 \\
UNION disjunction    &    51   &  0.8 \\
Optional (OPTIONAL)  &    41   &  0.7 \\
\midrule
Total & 6{,}136 & 100 \\
\bottomrule
\end{tabular}
\end{table}

\subsection{Natural-Language Question Characteristics}
\label{sec:nl_chars}

The 30{,}680 questions show systematic variation in length across the five register
points. Q1 is the shortest (13.8 words on average) because ontology class names and property
identifiers are concise tokens that avoid elaboration.
Q2 questions are the longest (16.2 words), as the systems-engineering register tends
to expand ontology class names into multi-word engineering descriptors (e.g.,
``zone air cooling temperature setpoint'' for \texttt{Zone\_Air\_Cooling\_Temperature\_Setpoint}).
Q3 questions are shorter (14.2 words) because the operational register strips schema vocabulary
and describes needs more abstractly. Q4 questions (15.5 words) are longer than Q1, reflecting
the cost of naming a specific instance identifier and framing a practical task around it. Q5 is
somewhat shorter than Q4 (14.6 words) because it removes both engineering vocabulary and any
equipment identifier, paraphrasing the information need in entirely generic terms.

Instance grounding, defined as the use of a specific named equipment identifier from the query's
validation-walk results, is achieved in 79\% of Q4 questions, as detected by automated
identifier-pattern matching. The Q4 prompt instructs the model to name a specific instance when
one is available from the sample rows. The remaining 21\% falls into two categories: queries
whose results contain only counts or vocabulary Uniform Resource
Identifiers (URIs) (aggregation and filtered patterns have no
named instances to embed), and queries whose instance names are common English words
(e.g., a zone labelled \texttt{room}) that are correctly grounded but indistinguishable from
generic class references by pattern matching. The 79\% figure is therefore a conservative lower
bound over the Q4 register. The Q5 prompt deliberately withholds named instance
identifiers, keeping the plain-English register free of equipment tags; Q5 grounding is
accordingly near zero by design.

Table~\ref{tab:example_queries} shows all five register points for a representative
linear-chain query. The spectrum has two perceptually distinct zones. The first is the
\textbf{vocabulary abstraction} zone (Q1--Q3): Q1 names ontology classes and property
identifiers verbatim (\texttt{brick:Start\_Stop\_Command}, \texttt{brick:isPointOf}); Q2
replaces schema vocabulary with engineering domain terminology (\textit{supply fan start/stop
command}); Q3 further simplifies to operational language, removing all schema vocabulary and
describing the information need generically. The second is the \textbf{instance-grounding
transition} (Q4--Q5): Q4 embeds a specific named equipment identifier drawn verbatim from
the query's validation-walk results (\texttt{SupplyFan\_AHU01}), framing the question as a
practical operational task around a concrete object. Q5 paraphrases the same need in plain
language without any equipment identifier. Questions for all remaining pattern types follow
the same five-register structure and are available in the dataset.
For contrast, a branching (Y-shape) Q1 question expresses a two-attribute co-occurrence:
\textit{``Which \texttt{brick:VAV} instances are linked via \texttt{brick:feeds} to a
\texttt{brick:HVAC\_Zone} and via \texttt{brick:hasPoint} to a
\texttt{brick:Zone\_Air\_Temperature\_Sensor}?''} -- qualitatively different in structure from any
single-path linear question.

The corpus employs 232 distinct Brick class and property terms across HVAC, electrical,
metering, spatial, and sensing subsystems; the five most-queried terms
(\texttt{hasPoint}, \texttt{VAV}, \texttt{feeds}, \texttt{HVAC\_Zone},
\texttt{Air\_Handler\_Unit}) reflect the air-side HVAC focus of the Brick ontology.

\begin{table}[tbp]
\centering
\small
\caption{Five register levels for a representative linear-chain query.}
\label{tab:example_queries}
\begin{spacing}{1.0}
\begin{tabular}{clp{9.5cm}}
\toprule
Q & Register & Question \\
\midrule
1 & Technical &
  Which \texttt{brick:Start\_Stop\_Command} instances are linked via
  \texttt{brick:isPointOf} to \texttt{brick:Supply\_Fan} entities? \\[3pt]
2 & Systems-engineering &
  Which supply fans are associated with which start/stop command points? \\[3pt]
3 & Operational &
  Which supply fans have corresponding start and stop controls? \\[3pt]
4 & Instance-grounded &
  Which start/stop command is mapped to \texttt{SupplyFan\_AHU01}? \\[3pt]
5 & Plain English &
  Which control command is used to turn the supply fan on and off? \\
\bottomrule
\end{tabular}
\end{spacing}
\end{table}

\section{Validation}
\label{sec:validation}

Build2SPARQL is validated along three dimensions. SPARQL correctness is guaranteed by
construction through pre-publication executability verification. Question quality is examined
per pattern type and corroborated by an independent two-rater human validation study.
Functional utility is demonstrated through a baseline NL-to-SPARQL
evaluation across three open-weight LLMs. The corpus is then compared against BuildingQA
across scale, pattern coverage, question quality, and overlap in pattern types.

\subsection{Executability Verification}
\label{sec:executability}

Every SPARQL query in the corpus is guaranteed executable by construction. The triple-walking
validation stage (Methodology section) discards any candidate pattern that returns no
result rows against its source building graph, so retention in the corpus implies executability.
No query enters the NL generation stage until its traversal path is confirmed to
exist with at least one matching instance in the actual graph.

The NL-to-SPARQL baseline evaluation below provides direct
corroboration on the held-out subset: ground-truth queries are re-executed against their source
KGs to obtain result sets for comparison, and all ground-truth queries on the 28 test buildings
executed successfully. Users wishing to independently verify the full corpus can obtain the source
building graphs from their original repositories (linked in the Zenodo deposit) and load them into
any SPARQL~1.1 compliant engine; the dataset metadata maps each
query record to its source building identifier.

\subsection{Per-Pattern-Type Quality Breakdown}
\label{sec:per_type}

Table~\ref{tab:per_type_quality} breaks down three question quality metrics by pattern type.
\textit{Lexical self-similarity} is the mean pairwise Jaccard similarity across the
$\binom{5}{2}$ question pairs per query (lower = more lexically diverse).
\textit{Distinct-opening rate} is the fraction of unique first tokens across the five questions
for a query, a proxy for syntactic variety (higher = more varied sentence openings).
\textit{Q4 instance-grounding rate} is the fraction of Q4 questions that embed a specific named
equipment identifier drawn from the validation-walk results.

Branching and optional queries score highest on distinct-opening rate (0.910 and 0.883
respectively), reflecting that multi-relation structures naturally require varied sentence
constructions to express joint or conditional relationships. Linear and
branching queries show the highest Q4 instance-grounding rates (81\% and 82\%), reflecting
that traversal queries typically surface concrete equipment identifiers in their result sets
(e.g., ``Which sensors are assigned to \texttt{VAV\_F1\_Z01}?''). Aggregation (COUNT) and
union queries show lower rates (52\% and 22\%) because count-returning queries and disjunctive
paths over broad equipment classes yield fewer named equipment instances in their sample results.
Optional queries show the lowest grounding rate (17\%), as presence/absence patterns rarely
return specific instance names from their result sets.

\begin{table}[tbp]
\centering
\small
\caption{Quality metrics by pattern type.}
\label{tab:per_type_quality}
\begin{tabular}{lrrrr}
\toprule
Pattern type & $n$ & Lex.\ sim. & Dist.\ open. & Q4 inst.\ (\%) \\
\midrule
Linear chain         & 5{,}009 & 0.170 & 0.714 & 81 \\
Branching (Y-shape)  &   815   & 0.156 & 0.910 & 82 \\
Aggregation (COUNT)  &   120   & 0.176 & 0.753 & 52 \\
UNION disjunction    &    51   & 0.243 & 0.600 & 22 \\
Optional (OPTIONAL)  &    41   & 0.176 & 0.883 & 17 \\
Attribute-filtered   &   100   & 0.198 & 0.677 & 56 \\
\midrule
Corpus total         & 6{,}136 & 0.170 & 0.740 & 79 \\
\bottomrule
\end{tabular}
\end{table}

\subsection{Human Validation of Natural-Language Questions}
\label{sec:human_validation}

The automated metrics above characterize surface diversity and instance grounding, but do not on
their own establish that the generated questions are semantically faithful to their paired queries,
natural, and operationally meaningful. To supply direct human evidence, two raters with
building-engineering backgrounds (Appendix~I) independently
coded a stratified sample of 300 natural-language questions (60 queries spanning the six pattern 
types, each with its five register variants) on three criteria drawn from the reviewers' terms: 
\emph{semantic fidelity} (does the question ask for what the query returns), \emph{naturalness} 
(does it read as fluent English), and \emph{operational plausibility} (would a practitioner 
realistically ask it). Marks were scored $\mathrm{Y}{=}1$, $\mathrm{B}{=}0.5$, $\mathrm{N}{=}0$; 
the full protocol, criteria definitions, and calibration procedure are given in Appendix~I.

Inter-rater agreement by Cohen's $\kappa$~\cite{cohen1960kappa} was $0.04$ (semantic fidelity),
$-0.03$ (naturalness), and $0.53$ (operational plausibility), with raw agreement of $95.0\%$,
$94.3\%$, and $89.0\%$, respectively. The near-zero $\kappa$ on the first two criteria is a
prevalence artifact: both raters assigned a passing mark to roughly $97\%$ of questions, so chance
agreement is already very high and $\kappa$ is deflated despite strong raw agreement. The
prevalence-adjusted, bias-adjusted kappa (PABAK)~\cite{byrt1993pabak} --- $0.90$, $0.89$, and $0.78$
for the three criteria --- is therefore also reported, as it better reflects reliability under these
skewed marginals. After adjudicating every disagreement to a consensus mark against the codebook,
pass rates were $98.8\%$ for semantic fidelity, $98.8\%$ for naturalness, and $84.0\%$ for
operational plausibility; the full breakdown by register and pattern type is reported in Appendix~I.
Semantic fidelity and naturalness exceed $97\%$ in every register --- including instance-grounded
(Q4) questions, which remain faithful ($97.5\%$) once judged on grounding correctness. Operational
plausibility is the lowest and most variable criterion, and its shortfall is concentrated in one
query type. Of the 60 sampled queries, 12 failed this criterion. Eleven of them are the 11
ASHRAE~223P connection-topology queries in the sample, which relate junctions, connection points,
and connectable components through connection relations such as \texttt{s223:cnx} and
\texttt{s223:hasConnectionPoint}. The twelfth pairs a Brick site with its buildings. Every other
sampled query passed in all five registers, including all Brick UNION, aggregation, and OPTIONAL
queries. The lower rates for UNION ($60.0\%$), aggregation ($73.3\%$), and OPTIONAL ($76.7\%$)
therefore reflect the share of connection-topology queries in those strata. Because ratings depended
mainly on the query and not on its phrasing, these rates rest on few independent queries (six UNION,
nine aggregation, and six OPTIONAL), and a single query shifts a rate by up to 17 percentage points.
Within the failing queries, the instance-grounded (Q4) question still passed in 11 of 12 cases,
because naming a specific component turns a wiring lookup into a concrete field check.

These results bear directly on the practical relevance of the corpus. Connection-topology queries
account for 879 of the 6{,}136 released queries ($14.3\%$), including 43 of the 51 UNION queries, 24
of the 41 OPTIONAL queries, and 38 of the 120 aggregation queries. Weighting the sample pass rates
of connection-topology ($21.8\%$) and other queries ($98.0\%$) by their shares of the full corpus
gives a corpus-level plausibility of about $87\%$. The same rates imply only about $34\%$ for the
UNION family, in which connection-topology queries predominate. Connection-topology queries remain
valid, executable queries that exercise the 223P connection model, which matters for systems that
must traverse 223P graphs. They are less representative of the questions operators ask in routine
operation. Users who target operator-facing question answering can identify them by their 223P
connection relations and evaluate them separately. Results on the UNION family, most of which
consists of such queries, should not be read as evidence of operator-facing performance. Together
these results corroborate the automated metrics with human judgment and directly address the
reviewers' request for evidence of question quality beyond automated measures.

\subsection{NL-to-SPARQL Baseline Evaluation}
\label{sec:eval}

Three open-weight LLMs are evaluated on the NL-to-SPARQL task across three prompting conditions:
zero-shot, 3-shot with randomly selected in-context examples~\cite{brown2020fewshot}, and 3-shot
with retrieved in-context examples~\cite{lewis2020rag,karpukhin2020dpr}. Retrieved examples are selected by cosine-similarity search over
\texttt{nomic-embed-text}~\cite{nussbaum2024nomic} sentence embeddings~\cite{reimers2019sbert} of the retrieval-pool \emph{questions}: the test question is
embedded, and the three retrieval-pool questions whose embeddings are most cosine-similar to it are
injected into the prompt together with their ground-truth SPARQL. The retrieval key is therefore the
natural-language question itself -- the only input available at inference time -- rather than any
field produced by the generation pipeline, so the setup matches a realistic deployment in which a
library of example question--SPARQL pairs is queried by an incoming user question.
The embedding index covers the 173 retrieval-pool buildings; models are evaluated on the 28
held-out test buildings.

Each prompt shares a common preamble: the Brick schema description for the target building
(drawn from the \texttt{description} field of each query record) followed by the natural-language
question. The zero-shot condition appends no examples. The 3-shot random condition appends three
question--SPARQL pairs drawn uniformly at random from the retrieval pool. The 3-shot retrieved
condition replaces those with the three retrieval-pool examples whose \emph{question}
embeddings are most cosine-similar to the test question, so the injected examples share
phrasing and ontology patterns with the query being translated. Listing~\ref{lst:eval_prompt} shows the
complete template; square-bracketed blocks appear only in few-shot conditions.

\begin{spacing}{1.0}
\begin{lstlisting}[caption={NL-to-SPARQL evaluation prompt template.},
                   label={lst:eval_prompt}]
-- System message --
You are a SPARQL expert for building automation knowledge graphs.
Return only the SPARQL SELECT query - no explanation, no markdown.

-- User message --
Building: {building_id} | Ontology: Brick
{building_description}
SPARQL prefixes: {prefix_block}

[Question: "{example_question}"   <- repeated 3x; omitted in zero-shot
 SPARQL:   {example_sparql}      ]

Question: "{question_text}"
Write a SPARQL SELECT query that answers this question.
\end{lstlisting}
\end{spacing}

For each question, the model generates a SPARQL query, evaluated on four metrics.
\textit{Syntax}: whether the generated query is syntactically valid SPARQL.
\textit{Exec}: whether it executes without error against the source building graph.
\textit{Exact Match F1}: whether the result set exactly matches the ground-truth result set
(strict set equality over result rows).
\textit{Row Matching F1}: result-set F1 at the row level; following BuildingQA~\cite{mulayim2025buildingqa},
letting $\hat{R}$ be the predicted result set and $R$ the ground-truth (each a set of result rows),
\begin{equation}
  P = \frac{|\hat{R} \cap R|}{|\hat{R}|},\quad
  \text{Rec} = \frac{|\hat{R} \cap R|}{|R|},\quad
  \text{Row Matching F1} = \frac{2P \cdot \text{Rec}}{P + \text{Rec}}
  \label{eq:f1}
\end{equation}
with Row Matching F1 $= 0$ when $|\hat{R}| = 0$ or $|R| = 0$;
Exact Match F1 is the special case Row Matching F1 $= 1$.
BuildingQA's two additional metrics -- Arity Matching F1 (variable-count match) and
Entity Set F1 (per-column unique-value F1, computed only on arity-pass rows) -- are included
in Table~\ref{tab:eval_main}.

\paragraph{Models} Three open-weight models are evaluated: Gemma~4~4B
(Ollama: \texttt{gemma4:e4b}), Llama~3.1~8B (\texttt{llama3.1:8b})~\cite{grattafiori2024llama3}, and Qwen~2.5~32B
(\texttt{qwen2.5:32b})~\cite{yang2024qwen25}. They were selected to span three distinct open-weight model families
(Google, Meta, Alibaba) and three parameter-count tiers (4B, 8B, 32B), enabling analysis of
both architectural diversity and scale effects on SPARQL generation quality. All three are
served locally via Ollama with no fine-tuning, matching the on-premise deployment context
realistic for building operators and ensuring full reproducibility without external API
dependencies.

\begin{table}[tbp]
\centering
\small
\caption{NL-to-SPARQL baseline results on the held-out test set. All models achieve 100\% syntactic validity.}
\label{tab:eval_main}
\begin{tabular}{llrrrrr}
\toprule
Model & Condition & Arity F1 (\%) & Exec (\%) & Exact Match F1 (\%) & Row F1 & Ent.\ Set F1 \\
\midrule
Gemma~4~4B & Zero-shot          & 68.1 & 91.0 &  3.8 & 0.039 & 0.390 \\
Gemma~4~4B & 3-shot (random)    & 74.1 & 93.2 & 25.2 & 0.257 & 0.754 \\
Gemma~4~4B & 3-shot (retrieved) & 87.4 & 89.4 & 60.9 & 0.615 & 0.958 \\
\midrule
Llama~3.1~8B & Zero-shot          & 66.7 & 90.6 &  0.2 & 0.016 & 0.214 \\
Llama~3.1~8B & 3-shot (random)    & 71.5 & 88.1 & 22.0 & 0.229 & 0.797 \\
Llama~3.1~8B & 3-shot (retrieved) & 91.7 & 88.1 & 56.0 & 0.575 & 0.907 \\
\midrule
Qwen~2.5~32B & Zero-shot          & 71.7 & 92.6 & 20.4 & 0.216 & 0.757 \\
Qwen~2.5~32B & 3-shot (random)    & 80.9 & 93.9 & 35.3 & 0.358 & 0.841 \\
Qwen~2.5~32B & 3-shot (retrieved) & 93.0 & 90.5 & 65.2 & 0.659 & 0.969 \\
\bottomrule
\end{tabular}
\end{table}

\paragraph{Results} Table~\ref{tab:eval_main} shows the main results. All models produce
syntactically valid SPARQL in 100\% of cases across all conditions. Execution success (88--94\%)
and Exact Match F1 (0.2--65\%) vary substantially across conditions.

The dominant finding is that retrieval quality determines performance: 3-shot retrieved
dramatically outperforms both zero-shot and random 3-shot across all three models, lifting
Exact Match F1 from under 21\% to 56--65\%, Row Matching F1 from under 0.22 to 0.58--0.66,
and Entity Set F1 from under 0.76 to 0.91--0.97.
Random 3-shot provides substantially more benefit than zero-shot when retrieved examples are
unavailable (22--35\% Exact Match F1 vs.\ 0.2--20\% zero-shot).

Table~\ref{tab:eval_by_level} and Figure~2 show the per-register
performance under 3-shot retrieved. Q1 questions (ontology terms explicit) are the easiest to
translate (85.1\% Exact Match F1 for Qwen), and Q2--Q3 follow the expected vocabulary-abstraction
gradient. Q4 (instance-grounded), however, is by far the worst-performing register (17--37\%
across models), substantially below even Q5 (plain English, 57--68\%). The cause is a
systematic evaluation mismatch rather than a simple model failure. A Q4 question names a
specific equipment instance -- for example, ``which start/stop command is mapped to
\texttt{SupplyFan\_AHU01}?'' -- and the model, reasonably, generates instance-specific SPARQL
that binds or filters to that IRI. The ground-truth query, however, is a generic structural
pattern that retrieves the relationship for \emph{all} supply fans with no instance filter;
the two result sets are therefore disjoint by construction, driving Exact Match F1 to near zero even
when the model has correctly identified the underlying structural pattern. Q5, which
deliberately withholds all instance identifiers, elicits structurally generic SPARQL and
scores substantially higher as a result. This gap points to a broader tension in
NL-to-SPARQL evaluation design: instance-grounded operational questions and generic structural
ground-truth queries measure different things, and the mismatch should be accounted for when
interpreting Q4 scores. Entity Set F1 by register (all conditions pooled) confirms the same
pattern: Q1--Q3 and Q5 score 0.83--0.88, while Q4 drops to 0.51. The register spectrum
therefore introduces two distinct difficulty axes: the vocabulary-abstraction gradient
(Q1$\to$Q3) and the instance-specificity trap at Q4---a separation of ontology-level reasoning
from building-instance grounding that has been proposed as a design principle for building
knowledge-graph question answering~\cite{jung2026bricktrace}.

\begin{table}[tbp]
\centering
\small
\caption{Exact Match F1 (\%) by register level, 3-shot retrieved.}
\label{tab:eval_by_level}
\begin{tabular}{llrrrrr}
\toprule
Model & Condition & Q1 & Q2 & Q3 & Q4 & Q5 \\
\midrule
Gemma~4~4B & Zero-shot          & 15.8 &  3.1 &  0.8 &  0.0 &  0.0 \\
Gemma~4~4B & 3-shot (random)    & 32.7 & 26.4 & 26.7 &  9.0 & 27.6 \\
Gemma~4~4B & 3-shot (retrieved) & 81.2 & 74.4 & 66.7 & 17.0 & 60.0 \\
\midrule
Llama~3.1~8B & Zero-shot          &  0.0 &  0.0 &  0.0 &  1.0 &  0.0 \\
Llama~3.1~8B & 3-shot (random)    & 27.7 & 22.5 & 21.7 & 14.0 & 27.6 \\
Llama~3.1~8B & 3-shot (retrieved) & 78.2 & 63.6 & 57.5 & 21.0 & 57.1 \\
\midrule
Qwen~2.5~32B & Zero-shot          & 26.7 & 24.8 & 19.2 &  9.0 & 22.9 \\
Qwen~2.5~32B & 3-shot (random)    & 41.6 & 40.3 & 40.0 & 22.0 & 37.1 \\
Qwen~2.5~32B & 3-shot (retrieved) & 85.1 & 69.0 & 65.8 & 37.0 & 67.6 \\
\bottomrule
\end{tabular}
\end{table}

\begin{figure}[tbp]
\centering
\includegraphics[width=0.65\linewidth]{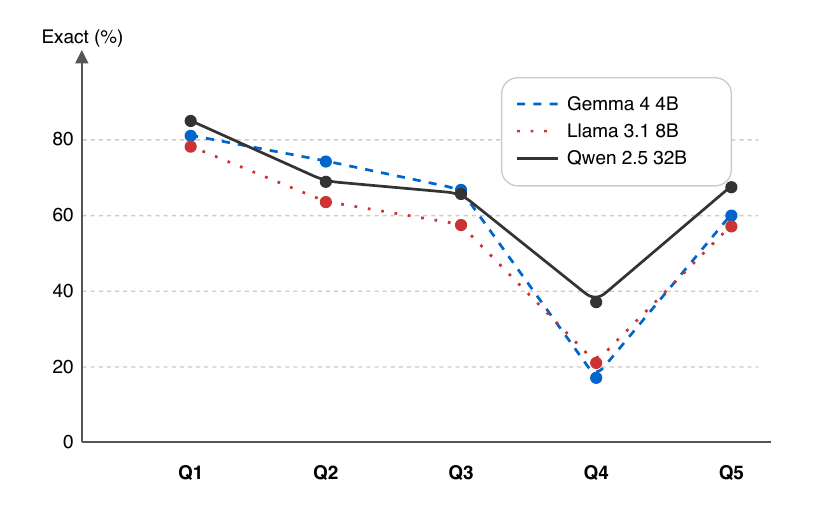}
\caption{Exact Match F1 (\%) by register, 3-shot retrieved. Q4 underperforms Q5 due to
instance-specificity mismatch.}
\label{fig:qlevel_gradient}
\end{figure}

Table~\ref{tab:eval_by_pattern} shows Exact Match F1 by SPARQL pattern type under 3-shot retrieved.
Linear chains dominate the 20-question-per-building evaluation sample (capped to bound runtime;
513 of 555 rows), so linear results are statistically robust. Non-linear types have small sample counts (5--15 rows each) because
their lower per-building frequency is diluted by the random sampling cap; these figures should
be interpreted as directional rather than precise. Among linear chains, retrieved-3 achieves
64--68\% Exact Match F1. UNION queries remain the hardest (9--18\%), consistent with the
two-branch structure being difficult to infer without an explicit structural example.

\begin{table}[tbp]
\centering
\small
\caption{Exact Match F1 (\%) by pattern type, 3-shot retrieved.}
\label{tab:eval_by_pattern}
\begin{tabular}{lrrrrrr}
\toprule
Model & Lin. & Brnch. & Union & Aggr. & Opt. & Filt. \\
\midrule
Gemma~4~4B    & 63.5 & 40.0 &  9.1 & 33.3 & 50.0 & 20.0 \\
Llama~3.1~8B  & 58.5 & 40.0 & 18.2 & 26.7 & 33.3 & 20.0 \\
Qwen~2.5~32B  & 67.6 & 40.0 & 18.2 & 26.7 & 66.7 & 60.0 \\
\bottomrule
\end{tabular}
\end{table}

\paragraph{Discussion} The high retrieved-3 performance reflects that the Brick ontology is
consistent across buildings: once a model sees the correct property chain from a retrieved
in-context example, it applies it to the test building. Splitting the 28 test buildings by type reveals an instructive asymmetry: the 15 hospital
BuildGraph buildings held entirely out of the retrieval pool achieve 67--74\% Exact Match F1, while the
13 randomly sampled real buildings score 43--55\%. The hospital buildings' higher performance
is consistent with their procedural origin -- BuildGraph-generated buildings share the same
generator's structural conventions as the synthetic retrieval-pool buildings, so retrieval locates
compatible examples despite the type difference. The real Mortar and Brick reference buildings
(43--55\%) are therefore the more meaningful indicator of practical deployment difficulty,
reflecting the heterogeneous naming conventions and irregular topologies of actual installed systems.

\paragraph{Retrieval-realism ablation} For every record, the writer stage produces a one-sentence
summary of what the query returns. Because this summary is a pipeline artifact that a deployed
system would not have, few-shot retrieval could in principle exploit it to inflate performance; an
ablation confirms that it does not. As noted above, retrieval embeds the natural-language
question, the only input available at inference; to rule out any hidden advantage from this
summary, a \emph{matched-modality oracle} is run that keys retrieval on the summary
at both index and query, giving it full access to that field. Table~\ref{tab:retrieval_ablation}
compares this oracle against the realistic question-keyed retrieval on a paired subset of eight real
Mortar buildings (same seed and sampled questions; only the retrieval key differs). The oracle never
outperforms the realistic configuration: for Gemma~4~4B ($n=120$) the two are statistically
indistinguishable (Exact Match F1 0.558 vs.\ 0.592; $\Delta = -3.3$ points), and for Llama~3.1~8B
(a smaller, timing-limited run, $n=45$) the oracle is clearly worse (0.511 vs.\ 0.667). A secondary
index-swap oracle---keying a question query against a summary index---degrades
Exact Match F1 from 0.65 to 0.33. Access to the pipeline field therefore confers no advantage,
confirming that the reported retrieval-augmented results reflect realistic, question-based retrieval.

\begin{table}[tbp]
\centering
\small
\caption{Retrieval-realism ablation: a matched-modality oracle (retrieval keyed on the pipeline's
query-result summary at both index and query) versus the realistic question-keyed
retrieval reported above, on a paired subset of eight real Mortar buildings.}
\label{tab:retrieval_ablation}
\begin{threeparttable}
\begin{tabular}{lrrrr}
\toprule
 & \multicolumn{2}{c}{Exact Match F1} & \multicolumn{2}{c}{Row F1} \\
\cmidrule(lr){2-3}\cmidrule(lr){4-5}
Model & Realistic & Oracle & Realistic & Oracle \\
\midrule
Gemma~4~4B ($n=120$)  & 0.592 & 0.558 & 0.598 & 0.566 \\
Llama~3.1~8B ($n=45$) & 0.667 & 0.511 & 0.682 & 0.519 \\
\bottomrule
\end{tabular}
\begin{tablenotes}\footnotesize
\item[] \textit{Note:} The oracle never beats the realistic configuration.
\end{tablenotes}
\end{threeparttable}
\end{table}

\subsection{Comparison with BuildingQA}
\label{sec:buildingqa_comparison}

\paragraph{Roles and pattern coverage} The two resources serve different roles. BuildingQA is a
small expert-authored evaluation set (27 queries, 188 NL questions, 4 buildings), and Build2SPARQL
is an automatically generated corpus (6{,}136 queries, 30{,}680 NL questions, 201 buildings)
(Table~\ref{tab:buildingqa_comparison}). The difference in counts reflects the generation method and
does not by itself indicate higher quality. Build2SPARQL implements six query-pattern families,
three of which appear in BuildingQA. Aggregation is absent from BuildingQA, and optional patterns
cannot be assessed from its 27 queries. Its authors attribute the aggregation gap to a practitioner
workflow convention (operators aggregate tabular exports rather than within SPARQL) rather than an
absent information need.

\begin{table}[h!]
\centering
\small
\caption{Build2SPARQL vs.\ BuildingQA.}
\label{tab:buildingqa_comparison}
\begin{tabular}{lrr}
\toprule
Metric & Build2SPARQL & BuildingQA \\
\midrule
\multicolumn{3}{l}{\textit{Scale and coverage}} \\
Buildings                              & 201      &  4 \\
Queries                                & 6{,}136  & 27 \\
NL questions                           & 30{,}680 & 188 \\
Pattern types                          & 6        & 3 \\
Mean BGP triples                       & 3.9      & 5.7 \\
\midrule
\multicolumn{3}{l}{\textit{Question quality}} \\
Lex.\ self-similarity (lower is better) & 0.170 & 0.238 \\
Distinct opening words / total          & 0.740 & 0.641 \\
Q4 instance-grounding rate (\%)         & 79    & 7 \\
Distinct ontology terms                 & 232   & 61 \\
\bottomrule
\end{tabular}
\end{table}

\paragraph{Question quality} Using the metrics defined in the Per-Pattern-Type Quality subsection above, plus
\textit{distinct ontology terms} (unique Brick class and property identifiers across all SPARQL
queries), Table~\ref{tab:buildingqa_comparison} presents the full quality comparison.

Build2SPARQL's lexical self-similarity (0.170) is lower than BuildingQA's (0.238), indicating that
the five-register prompt design maintains vocabulary diversity across levels. Generated corpora tend
to reuse common sentence frames as they grow, and the anti-repetition prompting is intended to
counteract this tendency. Distinct-opening-word variety is 0.740, compared with 0.641 for
BuildingQA. Because the two resources differ greatly in size and construction, these figures show
that diversity holds up in the generated corpus. They are not a ranking of the two resources.
The Q4 instance-grounding rate is 79\%, compared with 7\% in BuildingQA, reflecting
the Q4 prompt instruction to embed a specific instance name from validation-walk results when
one is available. The 79\% figure is a conservative lower bound: aggregation and union patterns
return counts or broad class sets with no named equipment instance to detect, so those queries
register no grounding even when the Q4 prompt was followed correctly. Build2SPARQL uses 232 distinct
ontology terms versus 61 in BuildingQA. This difference largely reflects the number of source
buildings (201 vs.\ 4), together with the multi-hop BFS enumeration, which surfaces terminal classes
reachable only through intermediate equipment relationships.
To verify that lexical diversity does not come at the cost of semantic accuracy, the two-rater human
validation (Human Validation subsection) found 98.8\% semantic fidelity across the register spectrum.

The one dimension where BuildingQA scores higher is mean basic graph pattern
(BGP) triple count (5.7 vs.\ 3.9),
reflecting that its 27 hand-authored queries target the multi-hop complexity of real operational tasks.

\paragraph{Overlap in pattern types} As established in the Related Work section, BuildingQA's
queries were written by domain experts for their own applications, without reference to the six
pattern types. They therefore provide an independent check on those types. Its 27 queries explicitly
include UNION disjunctions, multi-branch joins (branching Y-shape), and simple forward traversals
(linear chain); SPARQL aggregation (COUNT/GROUP~BY) is explicitly absent, with its authors
attributing the gap to a workflow convention -- operators aggregate on extracted tabular data rather
than within SPARQL -- rather than an absent information need~\cite{mulayim2025buildingqa}. Three of
the six pattern types are thus independently confirmed in practitioner-authored queries. The
remaining two types (optional-extension and attribute-filtered) cannot be assessed from the 27-query
sample. This overlap suggests that the six pattern types reflect practitioner information needs that
map to static graph queries, with the aggregation gap marking a boundary between what practitioners
currently encode in SPARQL and what they route through post-processing.

\subsection{Comparison with Automated and Prior Building-Domain Corpora}
\label{sec:automated_comparison}

The comparison with BuildingQA above concerns question quality against a hand-authored expert
benchmark. A complementary axis is structural coverage against the other automated and
semi-automated building-domain NL/SPARQL resources. Table~\ref{tab:automated_comparison} places
Build2SPARQL alongside BuildingQA~\cite{mulayim2025buildingqa}, the two smaller method-evaluation
efforts (DA-KGQA~\cite{mulayim2025zero} and OntoSage's~\cite{devmane2026ontosage} test protocol), and
the large template- and synthesis-based corpora (BuildingGPT~\cite{li2025buildinggpt},
BuildingGPT2~\cite{li2026buildinggpt2}, and OntoSage's training corpus).

Build2SPARQL is not the largest of these corpora: OntoSage reports roughly 120{,}000 synthetically
generated NL/SPARQL pairs, and BuildingGPT and BuildingGPT2 contain 18{,}555 distinct
template-instantiated queries, more than Build2SPARQL's 6{,}136. Its contribution is instead the
combination of four properties that none of the resources in Table~\ref{tab:automated_comparison}
combines. First, a broader set of structural query-pattern families -- branching joins,
\texttt{UNION} disjunctions, \texttt{OPTIONAL} extensions, grouped aggregation (\texttt{COUNT} with
\texttt{GROUP BY}), and unit/quantity attribute filtering, at one to five hops. Exclusion patterns
and post-aggregation filters remain outside this set. The template corpora cover linear one- to
two-hop retrieval, flat instance counting, and class-type filtering; OntoSage's released pairs do
not document these structural operators, and its aggregation and temporal reasoning are delegated to
downstream analytics microservices rather than expressed as SPARQL patterns. Second, multi-ontology
grounding: Build2SPARQL spans Brick and ASHRAE~223P, whereas BuildingGPT, BuildingGPT2, and OntoSage
are Brick-only. Third, source-building diversity: 201 buildings (real and synthetic) against 45,
three, and one. Fourth, register-differentiated phrasing: five questions per query across the
register spectrum rather than a single annotation.

Two qualifications keep the comparison honest. Executability is not unique to Build2SPARQL --
BuildingGPT, BuildingGPT2, and OntoSage all validate generated queries against an endpoint before
inclusion. And Build2SPARQL does not address the temporal and threshold reasoning that operational
practice also requires -- these lie outside the static-structure inventory and are discussed as
limitations in the Discussion section; OntoSage's analytics layer is one route to those needs.
Build2SPARQL's contribution therefore lies in breadth across query-pattern families, ontologies,
source buildings, and phrasing, released as a documented, executable benchmark.

\begin{table}[tbp]
\centering
\footnotesize
\setlength{\tabcolsep}{4.5pt}
\caption{Structural comparison of building-domain NL/SPARQL resources. Operator columns: Br
branching, Un \texttt{UNION}, Op \texttt{OPTIONAL}, Ag grouped aggregation, Fi unit/quantity filter; FT (in Construction) denotes
fine-tuning.
Marks: Y present, N absent, P partial, n.r.\ not reported. P denotes flat \texttt{COUNT} of a single
class (BuildingGPT/BuildingGPT2) or aggregation handled outside SPARQL (OntoSage).}
\label{tab:automated_comparison}
\begin{threeparttable}
\begin{tabular}{llrlrrccccc}
\toprule
Resource & Construction & Bldg. & Ontology & Queries & NL q. & Br & Un & Op & Ag & Fi \\
\midrule
BuildingQA\tnote{a}    & expert       & 4   & Brick, 223P & 27       & 188      & Y & Y & n.r. & N & n.r. \\
DA-KGQA                & method eval  & 1   & Brick       & 19       & 19       & N & N & N    & N & N \\
OntoSage\tnote{b}      & synthesis+FT & 3   & Brick       & \multicolumn{2}{c}{$\sim$120{,}000 pairs} & n.r. & n.r. & n.r. & P & n.r. \\
BuildingGPT            & 6 templates  & 45  & Brick       & 18{,}555 & 18{,}555 & N & N & N    & P & N \\
BuildingGPT2\tnote{c}  & templates+FT & 45  & Brick       & 18{,}555 & 18{,}555 & N & N & N    & P & N \\
\textbf{Build2SPARQL}  & KG-mined     & \textbf{201} & Brick, 223P & 6{,}136 & \textbf{30{,}680} & Y & Y & Y & Y & Y \\
\bottomrule
\end{tabular}
\begin{tablenotes}\footnotesize
\item[a] Question quality is compared separately in Table~\ref{tab:buildingqa_comparison}; three of
the six operator types appear, aggregation is explicitly absent, and \texttt{OPTIONAL}/filter cannot
be assessed from the 27-query sample.
\item[b] OntoSage reports $\sim$120{,}000 NL/SPARQL training pairs (50-pair test set); its released
pairs do not document these structural operators, and aggregation/temporal reasoning is handled by
analytics microservices, not SPARQL.
\item[c] BuildingGPT2 fine-tunes on 40 of the 45 buildings (17{,}019 pairs), holding out five.
\end{tablenotes}
\end{threeparttable}
\end{table}

\section{Discussion}
\label{sec:discussion}

\subsection{Research Applications}

The NL/SPARQL pairs produced by Build2SPARQL support three principal research applications.
\textbf{Text-to-SPARQL model development and evaluation:} pairs provide the scale and
held-out evaluation sets for building-domain SPARQL generation research. The building-level
split design (described in the Reference Evaluation Protocol subsection below) prevents equipment-identifier leakage, making
retrieval-pool/test separation trustworthy in a way random query-level splits cannot achieve.
\textbf{Building KG completeness assessment:} the mined pattern inventory reveals what a
KG currently exposes; absent classes and relationship types indicate
under-populated graph elements. Exclusion-pattern queries (\texttt{FILTER NOT EXISTS}) would make
this assessment direct instead of inferred from pattern absence.
\textbf{Cross-building benchmarking:} identical pattern templates applied uniformly across all
201 buildings enable controlled comparisons of queryability across building types, ontology
versions, and data quality levels -- a structured comparison manually authored benchmarks
cannot provide. Secondary applications include few-shot LLM prompting, SPARQL engine testing,
operator training materials, and grounding for LLM-based building assistants that generate operational
feedback and energy-saving recommendations~\cite{jung2026ecofeedback,jung2026cot}.

\subsection{Reference Evaluation Protocol}
\label{sec:evalprotocol}

\paragraph{Why building-level splits are required}
Any evaluation split must be defined at the \emph{building level}: partitioning at the query level
would leak building-specific equipment identifiers between the retrieval pool and test. Q4 and Q5 questions
embed verbatim instance names drawn from each building's validation-walk results (e.g.,
\texttt{SupplyFan\_AHU01}, \texttt{VAV\_F1\_Z01}); a model that sees those identifiers in the retrieval pool
and again in test is being evaluated on memorized vocabulary rather than structural generalization.
Users defining alternative splits should respect this building-level constraint.

\paragraph{Reference split design}
The reference split assigns 173 buildings to the retrieval pool and 28 to test (86/14 by building
count; 5{,}268 and 868 queries respectively). The test set has two components. First, all
15 BuildGraph hospital buildings are held out as a deliberate held-out type probe:
no hospital-type graph appears in the retrieval pool, testing type transfer to an unseen building
category. Second, 13 real buildings are randomly sampled into the test set -- 8 from the
Mortar pool and 5 from the Brick and ASHRAE~223P reference pool -- drawn with random
seed~42 and no further stratification. Table~\ref{tab:split_summary} lists the complete
test set by pool and building identifier. The companion file \texttt{split.json} in the
Zenodo release maps each of the 201 building identifiers to \texttt{``train''} (retrieval pool) or
\texttt{``test''}.

\begin{table}[h!]
\centering
\small
\caption{Reference evaluation split. Test buildings listed by pool; retrieval pool is all
  remaining 173 buildings.}
\label{tab:split_summary}
\begin{tabular}{lp{6.2cm}r}
\toprule
Pool & Test building identifiers & $n$ \\
\midrule
BuildGraph hospital &
  \texttt{BG\_hospital\_\{2004, 2013, pre1980\}}\newline
  \texttt{\_0000--0004} & 15 \\
Mortar &
  \texttt{bldg6, bldg10, bldg15, bldg16, bldg17,
  bldg22, bldg23, bldg25} & 8 \\
Brick / 223P reference &
  \texttt{apartment, IBAL, NIST-HPL,
  pnnl-bdg2-1, soda\_brick} & 5 \\
\midrule
\multicolumn{2}{l}{Total test} & 28 \\
\multicolumn{2}{l}{Retrieval pool (all other)} & 173 \\
\bottomrule
\end{tabular}
\end{table}

\subsection{Limitations}
\label{sec:limitations}

\textbf{Structural imbalance.} Linear chains constitute 81.6\% of the corpus. This reflects Brick topology rather than a generation bias, but it
can skew models trained on the raw distribution toward simple retrieval; the \texttt{pattern\_type}
field enables balanced or reweighted subsets, and per-pattern metrics should be reported.

\textbf{Top-down generation ceiling.}
All QA pairs are derived top-down: SPARQL queries are first mined from graph structure, then
natural-language questions are generated from those queries. Every pair is therefore expressible
within one of the six implemented structural templates. The benchmark measures whether a model
can identify the correct template and substitute the appropriate ontology class names; it does
not measure whether a model can handle questions whose SPARQL falls outside the template
inventory. Temporal queries (``has setpoint X changed in the last 24 hours?''), threshold
reasoning (``which zones consistently exceed their cooling setpoint?''), and cross-building
comparisons lie outside this inventory and are not represented in the evaluation. Practitioners
building production NL-to-SPARQL systems should treat the benchmark as a lower bound on
real-world difficulty rather than an upper bound.

\textbf{SPARQL fragment scope.} The corpus covers six families of SELECT queries, all built from
patterns present in the source graph: linear chains, branching joins, \texttt{UNION} disjunctions,
\texttt{COUNT} aggregations, \texttt{OPTIONAL} extensions, and attribute filters. Queries about
absence (\texttt{FILTER NOT EXISTS}, \texttt{MINUS}) and threshold filters on aggregates
(\texttt{HAVING}) are not generated. Absence queries require enumerating the complement set and
inverting the discovery logic, and \texttt{HAVING} predicates require domain-knowledge
thresholds that the pipeline does not define. Neither type appears in the 27 BuildingQA queries,
consistent with both being less common in small practitioner corpora; their absence does not
affect evaluation of the six currently implemented pattern types.

\textbf{Synthetic--real structural differences and generalizability.} 120 of the 201 buildings are
procedurally generated by BuildGraph~\cite{buildgraph2025}, covering 8 DOE commercial prototype
archetypes, and are clearly labeled in the dataset metadata; the pipeline's correctness guarantee is
graph-agnostic. The two populations differ structurally in ways visible in the corpus itself.
Synthetic graphs are regular: instances of the same archetype saturate the same pattern classes,
producing tightly clustered per-building query counts (408--460 per archetype)
and a lower branching rate. Real graphs are heterogeneous: query counts
range from a handful to several hundred, reflecting genuine variation in
graph density and system completeness, and they carry vendor-specific naming conventions, irregular
topologies, and a higher per-building branching rate as equipment
participates in more simultaneous relationships. These differences have a direct, measurable effect
on generalization: in the NL-to-SPARQL evaluation, the held-out synthetic
(BuildGraph hospital) buildings reach 67--74\% Exact Match F1, whereas the held-out real Mortar and
Brick reference buildings score 43--55\% -- a roughly 20-point gap that quantifies the additional
difficulty of installed systems and indicates that aggregate accuracy on the mixed corpus overstates
real-deployment performance. Two practices follow for researchers targeting real buildings: report
results on a real-only held-out set (the 81 real buildings are provided as a labeled subset), and
treat the synthetic buildings as a source of controlled structural diversity and coverage rather
than as a proxy for real-deployment difficulty.

\textbf{LLM-generated natural language.} The five NL questions per query are produced by a
LLM and may carry linguistic patterns or subtle biases from its pre-training
data. The pipeline's correctness guarantee applies only to the ground-truth SPARQL queries. A
two-rater human validation (Human Validation subsection) found high semantic fidelity and
naturalness on a stratified sample; nonetheless, the questions have not been authored or reviewed at
the scale of BuildingQA's interview-based construction, and a larger human rating effort would
further strengthen confidence in question quality.

\section{Conclusion}
\label{sec:conclusion}

Build2SPARQL was presented as a large-scale NL/SPARQL benchmark dataset for building KGs,
produced by a KG-grounded generation pipeline. The central design insight is the separation of correctness from linguistic
generation: SPARQL queries are produced entirely by graph-traversal code that mines structural patterns
directly from the source graph and validates each query through Python triple-walking, while LLMs
are used only to generate NL questions downstream of a verified query. The LLM
never writes SPARQL; the KG is therefore the correctness oracle and the resulting dataset is guaranteed
executable against the source buildings.

The pipeline implements six query-pattern families -- linear chains, Y-shape branches, UNION
disjunctions, COUNT aggregations, OPTIONAL extensions, and attribute-filtered chains -- defined over
the static structure of named-node KGs. Applied to 201 building KGs (81 real, 120 BuildGraph synthetic covering 8 DOE commercial prototype
archetypes), the pipeline produces 6{,}136 executable SPARQL queries and 30{,}680 NL questions
across five vocabulary registers spanning a technical-to-operational spectrum. Quality metrics confirm that the register-spectrum design achieves vocabulary diversity at
scale: mean pairwise lexical self-similarity is 0.170 across the five register levels, and
79\% of Q4 questions successfully embed a named equipment instance drawn from the
validation-walk results. A retrieval-augmented few-shot evaluation across three
open-weight LLMs demonstrates that retrieved examples
substantially raise accuracy over zero-shot baselines (Exact Match F1 of 56--65\%,
Row Matching F1 up to 0.66 for Qwen~2.5~32B with 3-shot retrieved), confirming that the
corpus supports few-shot NL-to-SPARQL research.

The comparison with BuildingQA provides qualified support for the six pattern types. Three of them
(linear chains, branching Y-shapes, and UNION disjunctions) appear independently in BuildingQA's
practitioner-derived queries. SPARQL aggregation is absent there, and its authors attribute the gap
to a workflow convention, in which operators aggregate extracted tabular data outside SPARQL, and
not to an absent information need.

Several directions remain open for extending the corpus and pipeline.
Two structural gaps in the current SPARQL inventory are the most direct targets for extension.
\texttt{FILTER NOT EXISTS} queries -- ``which zones have no temperature sensor?'' -- are the
highest-value unimplemented class; they invert the triple-walking discovery logic and would
elevate KG completeness assessment from inferring absence through missing patterns to direct,
query-by-query diagnostics. \texttt{HAVING} clauses offer a complementary extension: by
attaching threshold predicates to existing aggregation queries (e.g., ``air handling units (AHUs)
feeding more than
three VAVs''), they encode operational domain knowledge directly in SPARQL rather than pushing
it into post-processing. Implementing them requires a principled threshold-selection strategy,
either statistical or domain-knowledge-derived.

Beyond the current operator inventory, the pipeline's scope is bounded by the Brick ontology.
Building operations in practice draw on multiple co-existing schemas: ASHRAE~223P~\cite{ashrae223p}
governs Mechanical, Electrical, and Plumbing (MEP) system semantics, the Building Topology
Ontology (BOT)~\cite{rasmussen2021bot} captures spatial containment, and the Industry Foundation Classes in the Web Ontology Language (IFC-OWL) bridge
Building Information Modeling (BIM) authoring with the KG world. A unified multi-ontology corpus
would let model developers train and evaluate on the full range of building data an operator
actually encounters rather than Brick-only deployments. Extending the pipeline to these schemas
would require adapting pattern mining and triple-walking validation to each ontology's class
hierarchy and property vocabulary, but the pipeline's graph-agnostic correctness architecture
means the structural guarantee transfers directly.

The quality of the NL questions can be improved in parallel through a lightweight annotation
loop: human raters flag questions that fail phrasing variety or register fidelity, revised
prompts trigger targeted regeneration for those records, and pipeline version stamping isolates
which entries have been re-reviewed. This cycle can close the gap between automated generation
and the expert-authored queries of BuildingQA without requiring end-to-end human authorship at
corpus scale.

\section*{Appendix I: Human Validation Protocol}
\label{sec:appendix_validation}

Two reviewers requested direct human evidence that the LLM-generated natural-language (NL)
questions are semantically faithful to their paired SPARQL query, natural, and operationally
meaningful, beyond the automated metrics reported in the Validation section. This appendix
documents the protocol; the headline results are reported in the Human Validation subsection.

\paragraph{Sample and stratification.}
A stratified sample of 60 queries---and thus $60 \times 5 = 300$ NL questions, one per
register---was drawn from the released corpus. Queries were stratified across the six SPARQL pattern types in
proportion to their structural roles while guaranteeing a minimum cell size for the rarer types:
18 linear, 12 branching, 9 aggregation, 9 attribute-filtered, 6 UNION, and 6 optional. Every
sampled query carries its five register variants (Q1 Technical, Q2 Systems-engineering,
Q3 Operational, Q4 Instance-grounded, Q5 Plain English), so each register contributes exactly
60 questions.

\paragraph{Raters and calibration.}
Two raters coded the sample independently. One is a doctoral student in building engineering with
extensive experience in building energy data and building energy modeling, and the other is a
faculty member in building engineering. Both worked from a shared coding guide and a primer
covering the ontologies, query structure, and the register spectrum. To align interpretation with
the fixed codebook, the raters jointly coded the first three queries (15 questions) and reconciled
differences before coding the remaining items independently. Inter-rater agreement is reported in
full so that readers can judge the reliability of the ratings directly.

\paragraph{Criteria and scale.}
Each question was rated on three criteria drawn from the reviewers' own wording
(Table~\ref{tab:val_criteria}). Semantic fidelity (C1) and naturalness (C2) use a three-point scale
(Y, pass; B, borderline; N, fail); operational plausibility (C3), the most judgment-dependent
criterion, uses a binary Y/N scale. The SPARQL query is the authoritative description of what is
retrieved; a one-line natural-language gloss is provided only as a convenience, and raters were
instructed to defer to the SPARQL where the two appear to differ. A written note was required for
every B and every N.

\begin{table}[htbp]
\centering
\small
\caption{Human-validation criteria and rating scales.}
\label{tab:val_criteria}
\begin{tabular}{p{2.6cm}p{8.2cm}c}
\toprule
Criterion & Question the rater answers & Scale \\
\midrule
C1 Semantic fidelity & Does the question ask for what the paired SPARQL query actually
  returns---same entities, relationship, and conditions? & Y/B/N \\
C2 Naturalness & Does it read as fluent, grammatical English a practitioner might say? (For the
  Technical register, verbatim ontology tokens are expected and not penalized.) & Y/B/N \\
C3 Operational plausibility & Could a building operator, energy manager, or commissioning engineer
  realistically ask this in practice? & Y/N \\
\bottomrule
\end{tabular}
\end{table}

\paragraph{Scoring instance-grounded questions.}
The Instance-grounded register (Q4) names a single equipment instance drawn from the query's result
set by design, so naming one instance is not itself a fidelity defect. After the calibration phase,
the C1 rule for Q4 was therefore refined to judge \emph{grounding correctness}: a Q4 question is faithful
(Y) when the named instance is a genuine member of the query's result set and the relationship
matches; borderline (B) when it adds or alters a condition the query does not contain (for example,
asking only for the subset that is \emph{missing} an optional property); and a failure (N) when the
named instance is not in the result set or the relationship is wrong. This refinement was applied
symmetrically to both raters' Q4 codes and is the only change made to the coding scheme after
calibration.

\paragraph{Scoring, reliability, and adjudication.}
Marks were scored as Y${}=1$, B${}=0.5$, N${}=0$, and pass rates are reported per criterion, per
register, and per pattern type. Because the ratings are strongly Y-dominant, Cohen's $\kappa$ is
reported together with raw percentage agreement and the prevalence-adjusted, bias-adjusted kappa
(PABAK), which is more informative than $\kappa$ alone under skewed marginals. Every question on
which the two raters disagreed was resolved to a consensus mark by applying the codebook rules, and
final pass rates use consensus and adjudicated marks. The 65 disagreements (out of 900 marks)
concentrated in operational plausibility, where one rater tended to accept abstract connection- and
wiring-topology queries from the ASHRAE~223P models that the codebook classifies as non-operational;
these were resolved to the codebook classification.

Inter-rater agreement by Cohen's $\kappa$ was $0.04$ (C1), $-0.03$ (C2), and $0.53$ (C3), with raw
agreement of $95.0\%$, $94.3\%$, and $89.0\%$, respectively; the corresponding PABAK values, which
are more informative under the strong pass-prevalence, are $0.90$, $0.89$, and $0.78$. After
adjudication, pass rates were $98.8\%$ (semantic fidelity), $98.8\%$ (naturalness), and $84.0\%$
(operational plausibility). Table~\ref{tab:val_results} reports the full breakdown by register and
pattern type.

\begin{table}[htbp]
\centering
\small
\caption{Human-validation pass rates (\%; Y${}=1$, B${}=0.5$, N${}=0$) by register and by SPARQL
pattern type, after adjudication.}
\label{tab:val_results}
\begin{tabular}{lccc}
\toprule
Group & C1 Sem.\ fidelity & C2 Naturalness & C3 Oper.\ plaus. \\
\midrule
\multicolumn{4}{l}{\emph{By register}}\\
Q1 Technical          & 100.0 & 99.2 & 80.0 \\
Q2 Systems-engineering & 100.0 & 100.0 & 80.0 \\
Q3 Operational        & 99.2 & 98.3 & 80.0 \\
Q4 Instance-grounded  & 97.5 & 99.2 & 98.3 \\
Q5 Plain English      & 97.5 & 97.5 & 81.7 \\
\midrule
\multicolumn{4}{l}{\emph{By pattern type}}\\
Linear             & 99.4 & 98.3 & 85.6 \\
Branching          & 100.0 & 96.7 & 93.3 \\
UNION              & 100.0 & 100.0 & 60.0 \\
Aggregation        & 98.9 & 100.0 & 73.3 \\
Optional           & 91.7 & 100.0 & 76.7 \\
Attribute-filtered & 100.0 & 100.0 & 100.0 \\
\midrule
Overall            & 98.8 & 98.8 & 84.0 \\
\bottomrule
\end{tabular}
\end{table}

\paragraph{Availability.}
The full coding protocol, criteria, and analysis procedure are documented in this appendix; the
coding guide, rater primer, coded sheets, and analysis script are available from the author on
request.

\section*{Data Availability Statement}

The Build2SPARQL dataset is deposited at Zenodo (\url{https://zenodo.org/records/20451474}) and released
under a Creative Commons Attribution--NonCommercial 4.0 International (CC~BY-NC~4.0) license. The deposit provides
all NL/SPARQL pairs (one JavaScript Object Notation (JSON) file per building) and the reference
train/test split; the source building
graphs are not redistributed, but links to their original repositories are included.

\section*{Acknowledgments}

No funding was received in support of this study.

{\setstretch{1.0}


}

\end{document}